\documentclass[acmlarge]{acmart}

\usepackage{booktabs}
\usepackage{framed}
\usepackage{bm}
\usepackage[mathscr]{eucal}
\usepackage{amsmath}
\usepackage{txfonts}
\usepackage{multirow}
\usepackage{changepage}
\usepackage{url}
\usepackage{graphicx}

\usepackage{colortbl}
\usepackage{paralist}
\usepackage{xcolor}
\usepackage{makecell}
\usepackage{array}

\usepackage{color,soul}

\AtBeginDocument{%
  \providecommand\BibTeX{{%
    \normalfont B\kern-0.5em{\scshape i\kern-0.25em b}\kern-0.8em\TeX}}}

\begin{document}

\title{Towards Automated Factchecking: Developing an Annotation Schema and Benchmark for Consistent Automated Claim Detection}

\author{Lev Konstantinovskiy}
\email{lev.konst@gmail.com}
\affiliation{%
  \institution{Full Fact}
  \city{London, UK}
}

\author{Oliver Price}
\affiliation{%
  \institution{University of Warwick}
  \city{Coventry, UK}
}

\author{Mevan Babakar}
\affiliation{%
  \institution{Full Fact}
  \city{London, UK}
}

\author{Arkaitz Zubiaga}
\orcid{0000-0003-4583-3623}
\affiliation{%
  \institution{Queen Mary University of London}
  \city{London, UK}
}

\renewcommand{\shortauthors}{Konstantinovskiy et al.}

\begin{abstract}
 In an effort to assist factcheckers in the process of factchecking, we tackle the claim detection task, one of the necessary stages prior to determining the veracity of a claim. It consists of identifying the set of sentences, out of a long text, deemed capable of being factchecked. This paper is a collaborative work between Full Fact, an independent factchecking charity, and academic partners. Leveraging the expertise of professional factcheckers, we develop an annotation schema and a benchmark for automated claim detection that is more consistent across time, topics and annotators than previous approaches. Our annotation schema has been used to crowdsource the annotation of a dataset with sentences from UK political TV shows. We introduce an approach based on universal sentence representations to perform the classification, achieving an F1 score of 0.83, with over 5\% relative improvement over the state-of-the-art methods ClaimBuster and ClaimRank. The system was deployed in production and received positive user feedback.
\end{abstract}

% <concept_id>10010147.10010178.10010179</concept_id>
% <concept_desc>Computing methodologies~Natural language processing</concept_desc>
% <concept_id>10002951.10003317.10003347.10003356</concept_id>
% <concept_desc>Information systems~Clustering and classification</concept_desc>
% <concept_significance>300</concept_significance>
% <concept_id>10010405.10010497</concept_id>
% <concept_desc>Applied computing~Document management and text processing</concept_desc>
% <concept_significance>300</concept_significance>

% \ccsdesc[500]{Computing methodologies~Natural language processing}
% \ccsdesc[300]{Information systems~Clustering and classification}
% \ccsdesc[300]{Applied computing~Document management and text processing}


\maketitle

\section{Introduction}
\label{sec:introduction}

Misinformation has recently become more central in public discourse \cite{shu2017fake,zubiaga2018detection,ciampaglia2018digital}. As a consequence, interest has increased in the scientific community to further natural language processing (NLP) approaches that can help alleviate the burdensome and time-consuming human activity of factchecking \cite{vlachos2014fact, thorne2018automated}. Factchecking is known as the task of producing an informed assessment of the veracity of a claim \cite{graves2016understanding,gravesfactsheet}. The main mission of factcheckers is to give citizens information to make political choices, improve the quality of public political discourse and to hold politicians accountable \cite{graves2016understanding}. Misinformation and misperceptions can undermine this goal \cite{fridkin2015liar,flynn2017nature}.

However, there is a very small number of factchecking organisations in the world, about 160,\footnote{\url{https://reporterslab.org/fact-checking/}} compared to the volume of media items produced daily. The speed at which information flows online means there is less time to verify the claims made and myths spread further before being factchecked, if they are factchecked at all. Automating any parts of the factchecking process could cut down the time it takes to respond to a claim. It could also protect human factchecker's time to work on the more complicated checks that need careful human judgement.

When considered inside a factchecking organisation, the factchecking process consists of a series of tasks, which \cite{babakar2016state} defined as a process consisting of four tasks: (1) monitor media, i.e. capturing content such as articles, videos, images, (2) detect claims, i.e. spotting when an item contains a checkable claim, (3) check claims, i.e. doing the research to check and verify that claim, and (4) publish, i.e. creating a piece of content that encapsulates the results of the check.

The vast majority of the scientific research has focused on the determination of veracity \cite{huynh2018towards,thorne2018fever,zubiaga2018learning}, or checking claims; despite often having been given different names in the scientific literature, such as fake news detection \cite{wang2017liar,long2017fake,zhang2019overview} or arguably rumour detection \cite{yang2012automatic,ma2015detect,kwon2017rumor,alkhodair2019detecting}. 

The component checking the veracity of a claim cannot function in isolation, it needs a list of claims to check as an input. The majority of previous work on factchecking has started from a readily available list of claims and has omitted the previous steps of monitoring and spotting in the factchecking pipeline \cite{vlachos2014fact,huynh2018towards}. This act of spotting claims within a corpus is known as claim detection i.e. monitoring news sources and identifying if a sentence constitutes a claim that could be factchecked \cite{hassan2017claimbuster,Jaradat2018ClaimRankDC}. Monitoring and spotting within a factchecking organisation is a time consuming semi-manual task, which is inevitably limited by the resources available. Automating the process of claim detection could mean that factcheckers can monitor a greater set of media, extract the claims made, and hopefully make smarter choices about what the most valuable items to be checking that day are. If deployed in a live factchecking situation, it could also help separate out claims made in real-time during a ministerial speech, for example. This could help factcheckers quickly skim transcripts when time is limited. Despite the importance of this component in the factchecking pipeline, automation is still in its infancy and there is a dearth of scientific literature.

Previous works on claim detection \cite{Gencheva2017ACA,hassan2017claimbuster} rely on definitions of claims that incorporate the concept of check-worthiness and/or importance. Our objective is instead to avoid these subjective concepts to come up with an objective way of determining what constitutes a claim that is checkable; rather than being important or worthy of checking which is bound to subjective interpretation. This paper describes the iterative process we followed together with factcheckers to come with up an annotation schema that would effectively capture claims and non-claims. This annotation schema avoids factors that can be affected by personal biases, such as importance, in the manual annotation to produce an objective outcome. Following this annotation schema through a crowdsourcing methodology, we generated a dataset of 5,571 sentences labelled as claims or non-claims. Further, we set out to present the development of the first stage in the automated factchecking pipeline. It constitutes the first automated claim detection system developed by an independent factchecking charity, Full Fact,\footnote{http://fullfact.org/}
along with academic partners. The main contributions of our work are as follows:

\begin{itemize}
 \item We introduce the first annotation schema for claim detection, iteratively developed by experts at Full Fact, comprising 7 different labels.
 \item We describe a crowdsourcing methodology that enabled us to collect a dataset with 5,571 sentences labelled according to this schema.
 \item We develop a claim detection system that leverages universal sentence representations, as opposed to previous work that was limited to word-level representations. Our experiments show that our claim detection system outperforms the state-of-the-art claim detection systems, ClaimBuster and ClaimRank.
 \item With the annotation schema, crowdsourcing methodology and task definition, we set forth a benchmark methodology for further development of claim detection systems.
\end{itemize}

\section{Related Work}
\label{sec:related-work}

Most work around automatically factchecking claims has focused on the later stages of determining the veracity of claims, usually by building knowledge graphs out of knowledge bases, such as Wikidata \cite{wu2014toward,ciampaglia2015computational,shiralkar2017computational,shi2016discriminative,cao2018towards}. Less work has been documented on the preceding stage of claim detection.

One of the best-known approaches to claim detection is ClaimBuster \cite{hassan2017claimbuster}. They collected a large annotated corpus of televised debates in the USA. Their model combines Term Frequency-Inverse Document Frequency (TF-IDF), part-of-speech (POS) tags and named entity recognition (NER) features on an Support Vector Machines (SVM) classifier and produces a score of how important a claim is to factcheck. This has the caveat of then having to choose a cut-off score to determine the claims that will be considered worthy or important enough for factchecking. Our approach is instead to define an annotation schema that is binary, determining checkable claims rather than check-worthy claims, and is built on several types of claims. It is a simpler fit for the use case in this factchecking pipeline, i.e. in a live stream of subtitles we are unable to know in advance which sentence will make it to the top ranking until the end of the entire programme. 

In \cite{hanto2018towards} annotations were collected using ClaimBuster-inspired annotation guidance from volunteers together with their age, gender and education. With possible labels being verifiable check-worthy (VCW), verifiable not check-worthy (VNCW), and not verifiable (NV), they obtained 2,100 labelled sentences (reduced to 264 high-quality labelled sentences). They find that annotators with a natural sciences background agreed internally about what constitutes a check-worthy claim, whereas those with humanities, medicine, and ontology backgrounds saw more internal disagreement on check-worthiness. Furthermore, using 35 labelled control claims to test annotator skill, they find that the age group 40-49 obtains a higher average score, and label more claims, than that of 30-39, which correspondingly scores higher than age group 20-29.

Another recent approach to claim detection is ClaimRank \cite{Gencheva2017ACA}. They compiled a dataset by taking the outcome of factchecking a political debate, published by 9 organisations simultaneously. Models were created to predict if the claim would be highlighted by at least one or by a specific organisation. The modelling is done with a large variety of features from both the individual sentence and the wider context around it. A subsequent version of the dataset \cite{Jaradat2018ClaimRankDC,clef2018checkthat} includes a larger set of sentences in two languages, English and Arabic. These datasets are similar to ours in order of magnitude, however use a different definition of claim, as is the case with others tackling the determination of check-worthiness of claims \cite{hassan2015detecting,patwari2017tathya,barron2018overview,zuo2018hybrid}. We further elaborate on this in the next subsection.

A slightly different approach to claim detection is that of context-dependent claim detection (CDCD) \cite{levy2014context}. This study proposes identifying claims \textit{given a specific context}. Articles relevant to a topic are used to detect claims on that topic.

Another piece of work worth mentioning is the development of the FEVER (Fact Extraction and VERification) dataset \cite{thorne2018fever}. Whilst this is primarily aimed at work regarding claim veracity, the mere presence of a vast quantity of claims in the dataset allow it to be extended for claim detection in the future.

\subsection{Previous Attempts at Defining Claims}

There is a body of work on claim detection that has not formalised the definition of a claim e.g. \cite{Gencheva2017ACA,Jaradat2018ClaimRankDC}. Instead it directly relies on what has been identified by external organisations. The lack of a formal definition prevents others from replicating or extending their work. These studies used claims identified by 9 organisations in a political speech as a proxy. The annotations were sourced from publicly available online articles. This is different from our approach where we crowdsourced annotations following our definition of the task. The authors in \cite{Gencheva2017ACA} acknowledge this limitation, which led to high number of false positives in their experiments. For example, an article consisting of a debate transcript with editorial comments will not highlight repeated instances of claims. This creates inconsistent annotation -- during a TV debate, popular claims are discussed on repeated occasions. This had to be re-annotated by researchers in \cite{clef2018checkthat}. Another caveat is that only 3 of the 9 annotator organisations contributing to those online articles sign up to be neutral and transparent in their selection of claims as verified signatories of the International Factchecking Network's Code of Principles.\footnote{\url{https://perma.cc/BM43-SJ4N}}

ClaimBuster \cite{hassan2017claimbuster} provides a definition of a claim which revolves around the question: ``Will the general public be interested in knowing whether this sentence is true or false?''. Claims are considered to be those sentences for which the answer to this question is yes. Their aim was for anyone to be able to feed in a source, e.g. a political speech, and for the system to produce a list of claims ranked by importance, which could directly feed into the editorial process. This definition of a claim includes the judgement of `importance' which we avoid in our work. We believe it is an editorial judgement best left to factcheckers. ClaimBuster annotators were journalists, students and professors. Annotations that agreed with the authors of that study were selected to ensure good agreement and shared understanding of the assumptions. Researchers from the ClaimBuster team also defined an annotation schema called PolitiTax,\footnote{``PolitiTax A Taxonomy of Political Claims'' by IDIR Lab \url{https://perma.cc/4RQF-FCPV}} a taxonomy of political claims which we considered. However the categories were not useful for the downstream task of checking the veracity of the claim by routing it to the right dataset or team at Full Fact, in part due to the level of granularity in the taxonomy and in part because the team at Full Fact is split across topics.

There was also a taxonomy defined by factcheckers during the HeroX factchecking challenge \cite{herox2016francis}, which is less granular than PolitiTax. It has four claim types - numerical, political stance, quotes, objects. During this work we discovered that the latter three categories are rare and intersect with others, so we did not use them in our schema. 

\section{Dataset}
\label{sec:dataset}

\subsection{Our Claim Definition and Process}
\label{ssec:claim_def}

 Writing the annotation guidance was a long process. Full Fact's formal definition of a claim during the 2015 UK election, was ``an assertion about the world that can be checked''. Media monitoring volunteers were encouraged to ask a factchecker if they had doubts on whether something was assessable. We worked on codifying some of this thinking in conversations with the factcheckers. However, as we captured more and more claims, this definition proved insufficient. We wanted to understand whether a claim could be better defined by breaking it down into sub-categories for better consistency across time, topics and annotators. We opted for defining a typology that would capture the different types of claims, which would be more comprehensive than the previous short definition. Asking annotators to identify the category that a sentence belongs to would encourage more critical thinking. Likewise, choosing the right category for a sentence would significantly reduce the personal bias with respect to judging whether it is a claim, which is bound to personal judgement. Annotators would choose the category pertaining to a sentence, and we would then simplify the schema by mapping those types to binary labels, claim or not a claim.

We chose to decouple the importance of the claim from the claim itself. We felt that importance was heavily subjective, reliant on context and best left to factcheckers. Importance is a subtle, and forever changing feature. Even though the most ``important'' issues in the view of the UK public are often about the economy, immigration and health, their relative positions change\footnote{``Ipsos MORI Issues Index: 2017 in review'' \url{https://perma.cc/9SMV-CQR8}}. In some cases new issues become important, e.g. in the UK, the importance of claims about the EU increased significantly after the 2016 EU referendum. \footnote{``July 2016 Economist/Ipsos MORI Issues Index'' \url{https://perma.cc/DPA5-4XV5} }

We also chose to decouple the topic from the definition. By making our definition descriptive of the claim and not, by proxy, the topic, we would have a more consistent final dataset. In some cases the selection of topics is an inherently political choice, e.g. it varies across the population whether ``drugs'' relate to the topic of ``crime'' or ``health''. This kind of classification was avoided. 

To come up with the schema that would capture what constitutes a claim we followed an iterative process. In the first step, factcheckers identified sentences that were definitely not a claim. They iterated on potential rules and found examples that broke them. They also identified some constraints, for example, a claim needs to be checkable with more readily available evidence, which means that a personal claim like ``I woke up at 7am today'' is not a claim capable of being checked. We were most concretely able to exclude claims based on an individual's personal experience, as more often than not they were un-checkable. This is similar to `verifiable experiential' statements \cite{park2014identifying}.

We went through several versions of the guidance with different taxonomies. They were trialled within Full Fact, and then two versions with external volunteers. The first version applied the 2015 thinking and was a binary accept/reject classification task, accompanied by a guidance. It listed several types of qualities of claims and non-claims. Claims, for example, may be explicit, implicit, or trivial. Non-claims in this version were formed of personal experience and opinion. We decided against these categories in the end as they sometimes involve explicit judgements from our annotators -- these choices can sometimes be highly political. For example, in the case of \textit{``The EU is made up of 27 }[instead of 28]\textit{ countries''} or \textit{``The NHS is there for everyone''} some annotators could classify them as trivial while others might consider them explicit legal claims. The implicit/explicit categories were also removed, because whether the claim is implicit or explicit is not important for the next downstream task in the factchecking process after claim detection.

For the second version, we looked at Full Fact's factchecks. They mostly covered statistical claims. We also identified claims around current laws or rules of operation and correlation/causation claims e.g. ``there's no clear correlation between prisons' performance ratings and whether they're publicly-run or contracted out to the private sector.''. This became the basis of our claim categories. Merging these categories and removing personal experience was deemed to be a good proxy for claims. There were many other types of claims that we identified, such as definitions, voting records, and expressions of support. We limited our categories to 7 to make the task realistic for annotators. We also wanted to minimise the overlap between categories to make the task single-choice.

\subsection{Annotation Guidance}

Our annotation schema is the first to be created with a factchecking organisation. It comprises 7 categories, only one of which can be assigned to each sentence. Annotators were given definitions and examples of the 7 categories, including the more detailed breakdown shown in Table \ref{tab:category-breakdown}:

\begin{itemize}
    \item \textbf{Personal experience}. Claims that aren't capable of being checked using publicly-available information, e.g. ``I can't save for a deposit.''
    \item \textbf{Quantity in the past or present}. Current value of something e.g. ``1 in 4 wait longer than 6 weeks to be seen by a doctor.'' Changing quantity, e.g. ``The Coalition Government has created 1,000 jobs for every day it's been in office.'' Comparison, e.g. ``Free schools are outperforming state schools.''. Ranking, e.g. ``The UK's the largest importer from the Eurozone.''
    \item \textbf{Correlation or causation}, Correlation e.g. ``GCSEs are a better predictor than AS if a student will get a good degree.'' Causation, e.g. ``Tetanus vaccine causes infertility.'' Absence of a link, e.g. ``Grammar schools don't aid social mobility.''
    \item \textbf{Current laws or rules of operation}, Declarative sentences, which generally have the word "must" or legal terms, e.g. ``The UK allows a single adult to care for fewer children than other European countries.'' Procedures of public institutions, e.g. ``Local decisions about commissioning services are now taken by organisations that are led by clinicians.'' Rules and changes, e.g. ``EU residents cannot claim Jobseeker's Allowance if they have been in the country for 6 months and have not been able to find work.''
    \item \textbf{Prediction}, Hypothetical claims about the future e.g. ``Indeed, the IFS says that school funding will have fallen by 5\% in real terms by 2019 as a result of government policies.''
    \item \textbf{Other type of claim}, Voting records e.g ``You voted to leave, didn't you?'' Public Opinion e.g ``Public satisfaction with the NHS in Wales is lower than it is in England.'' Support e.g. ``The party promised free childcare'' Definitions, e.g. ``Illegal killing of people is what's known as murder.'' Any other sentence that you think is a claim.
    \item \textbf{Not a claim}, These are sentences that don't fall into any categories and aren't claims. e.g. ``What do you think?.'', ``Questions to the Prime Minister!''
\end{itemize}

\begin{table}
    \footnotesize
    \centering
    \begin{tabular}{|l|l|c|l|}
    \hline 
        Category & Subcategory & Counts & Example \\
        \hline
        \hline
        Not a claim & & 54.8\% & ``Give it all to them, I really don't mind.'' \\ 
        \hline
        \multirow{7}{*}{Other}
        & Other other & 10.4\%* & \makecell{``Molly gives so much of who she is away \\ throughout the film.''} \\
        \cline{2-4}
        & Support/policy & 5.5\%* & ``He has advocated for a junk food tax.'' \\
        \cline{2-4}
        & Quote & 4.7\%* & \makecell{``The Brexit secretary said he would guarantee \\ free movement of bankers.''} \\ 
        \cline{2-4}
         & Trivial claim & 1.6\%* & \makecell{``It was a close call.''} \\ 
        \cline{2-4}
         & Voting record & 0.7\%* & \makecell{``She just lost a parliamentary vote.''} \\
        \cline{2-4} 
         & Public opinion & 0.4\%* & \makecell{``A poll showed that most people who voted \\ Brexit were concerned with immigration.''} \\
        \cline{2-4}
         & Definition & 0.0\%* & \makecell{``The unemployed are only those actively \\ looking for work.''} \\
        \hline 
        \multirow{4}{*}{Quantity} & \makecell{Current value} & 9.9\% & \multirow{4}{*}{\makecell{``1 in 4 people wait longer than 6 weeks \\ to see a doctor.''}} \\
        \cline{2-2}
         & \makecell{Changing quantity} & & \\ 
        \cline{2-2}
         & \makecell{Comparison} & & \\
        \cline{2-2}
         & \makecell{Ranking} & & \\
        \hline 
        \multirow{2}{*}{Prediction} & \makecell{Hypothetical \\ statements} & 4.4\% & \makecell{``The IFS says that school funding will \\ have fallen by 5\% by 2019.''} \\
        \cline{2-2}
         & \makecell{Claims about \\ the future} & & \\
        \hline 
        \makecell{Personal \\experience} & \makecell{Uncheckable} & 3.0\% & ``I can't save for a deposit'' \\ 
        \hline 
        \multirow{3}{*}{\makecell{Correlation/ \\causation}} & \makecell{Correlation} & 2.6\% & ``Tetanus vaccine causes infertility'' \\ 
        \cline{2-2}
         & \makecell{Causation} & & \\
        \cline{2-2}
         & \makecell{Absence of \\ a link} & & \\
        \hline 
        \multirow{2}{*}{\makecell{Laws/rules of \\operation}} & \makecell{Public institution-\\al procedures} & 1.9\% & \multirow{2}{*}{\makecell{``The UK allows a single adult to care \\ for fewer children than other European \\ countries.''}} \\
        \cline{2-2}
         & \makecell{Rules/rule \\ changes} & & \\
    \hline
    \end{tabular}
    \caption{Breakdown of the 4,080 sentences with majority agreement. *The proportions for `Other' sub-categories are taken from a random sample of 160 claims labelled as `Other'.}
    \label{tab:category-breakdown}
\end{table}

These categories have proven to broadly cover sentences from political TV shows that Full Fact has encountered over several years. Categories have different levels of occurrence (see Table \ref{tab:category-breakdown}). As previously found \cite{Jaradat2018ClaimRankDC,hanto2018towards,hassan2017claimbuster}, ``Not a claim'' is the most popular category, amounting to about 55\% of the annotations. 

``Other'' is  the second largest category with 952 instances, 23\% of the whole. It can be broken down into claims that are less well-defined, with formal sub-categories being: `Definitions', `Voting records', `Public opinion', `Trivial claim', `Support', `Quote', `Other other'.\footnote{Note the `Other other' category is needed as sentence types previously unseen will inevitably emerge, however it is only intended for sentences that do not belong to any other category.} We amalgamate these because they are likely to overlap and we wanted our annotators to only select one option. For example, ``She said she voted to keep free school meals.'' is both a quote and a voting record. Furthermore, high granularity of categories allows one to perpetually think of rarer categories and a high number of categories slows down annotation unnecessarily. To verify if our level of granularity was correct, we split a sample of 160 sentences in `Other' into sub-categories. The vast majority are in the `Other other' category (see Table \ref{tab:category-breakdown}), supporting our chosen level of granularity.

\subsection{Crowdsourced Annotation}

The annotations were done by 80 volunteers recruited through Full Fact's newsletter -- this meant that volunteers were keen on factchecking. 28,100 annotations were collected for a set of 6,304 sentences extracted from subtitles of four UK political TV shows, 14 episodes in total. TV subtitles were chosen because 69\% of the UK population get their news from TV.\footnote{``News consumption in the UK:2016'' by Ofcom \url{https://perma.cc/5FDK-BRHD}}

The software used for collecting annotations was Prodigy\footnote{\url{https://prodi.gy/}},\footnote{https://fullfact.org/blog/2018/feb/how-we-customised-prodigy-ai/}
a self-hosted annotation platform. It was customised to support multiple annotators with a login and password screen where each user would enter their credentials. Sentences were shown in random order. The preceding two sentences were also shown on the screen to provide context and assist with potential co-references. Once a sentence was annotated 5 times by different annotators, it was not shown again.

Annotators were encouraged to contact us for any clarifications needed, with thoughtful questions such as: \textit{``Where it appears that a claim is dressed up as rhetorical question, should we classify it as a claim? e.g. `Why should unelected officials in Brussels make rules to stop bananas being sold in bunches of more than 2 or 3?'''} To answer this, questions are classified as the claims that they implicitly contain.

\begin{figure}[htb]
\centering
\includegraphics[width=7cm,height=8cm]{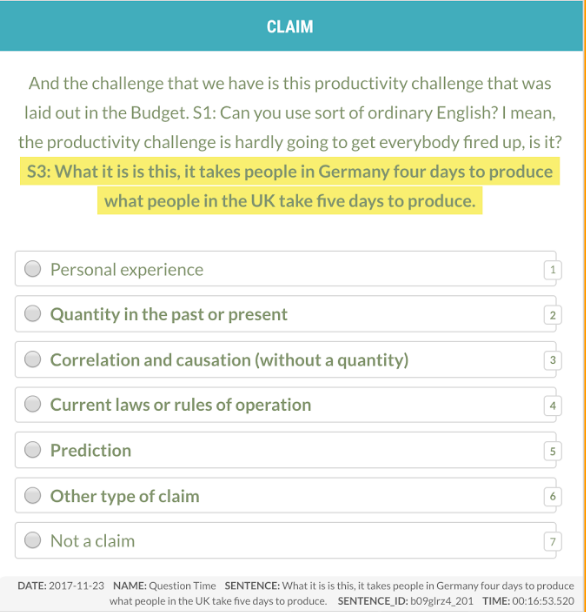}
  \caption{Annotation UI in Prodigy}
  \label{fig:annotation}
\end{figure}

\subsection{Agreement}
\label{ssec:agreement}

At the level of all the 7 granular categories the inter-annotator agreement is moderate, with a Krippendorff's alpha \cite{krippendorff1980reliability} of 0.46. However, we attain higher values of alpha of 0.70 and 0.53 when we do the mapping of annotations into the binary claim/non-claim annotation task, following either of the two methods shown in Table \ref{tab:reformulations}.

\begin{table}
  \centering
  \begin{tabular}{|c|c|c|c|c|c|c|}
    \cline{1-2}
    Qu & \cellcolor{green!50}12 & \multicolumn{5}{c}{} \\
    \cline{1-3}
    Corr & \cellcolor{green!50}10 & \cellcolor{green!50}10 & \multicolumn{4}{c}{} \\
    \cline{1-4}
    Law & \cellcolor{green!85}2  & \cellcolor{green!40}19 & \cellcolor{green!50}11 & \multicolumn{3}{c}{} \\
    \cline{1-5}
    Pred & \cellcolor{green!85}3  & \cellcolor{green!15}42 & \cellcolor{green!25}27 & \cellcolor{green!25}25 & \multicolumn{2}{c}{} \\
    \cline{1-6}
    Other & \cellcolor{yellow!30}50 & \cellcolor{yellow!60}102& \cellcolor{yellow!75}129& \cellcolor{yellow!40}87 & \cellcolor{yellow!40}87 & \multicolumn{1}{c}{} \\
    \hline
    Not & \cellcolor{yellow!60}114& \cellcolor{yellow!30}58 & \cellcolor{yellow!50}90 & \cellcolor{yellow!35}69 & \cellcolor{yellow!60}103& \cellcolor{red!70}668 \\%& \\
    \hline
     & Pers  & Qu  & Corr  & Law  & Pred  & Other \\% & 7 \\
    \hline
  \end{tabular}
  \caption{Annotation disagreements. The most prominent disagreement is between ``Other claim'' and ``Not a claim''. The labels are shortened versions of those in Figure \ref{fig:annotation} due to space limitations. (Qu: quantity, Corr: correlation and causation, Pred: predictions, Pers: personal experience.)}
  \label{tab:disagreements}
\end{table}

Most of the disagreement was between ``Not a claim'' and ``Other claim''. This showed that it is hard to define the boundary and explicitly list all kinds of claims as we saw in the process of creating the annotation guidance. The disagreements across all sentence types can be seen in Table \ref{tab:disagreements}.

\begin{table}[t]
  \centering
  \begin{tabular}{|c|c|c|c|c|c|}
    \hline
    \textbf{Claim} & \textbf{Non-claim} & \textbf{Omitted} & \textbf{$\alpha$} & N \\
    \hline
    2 & 3, 4, 6, 7 & 1, 5 & 0.70 & 6,095 \\
    \hline
    2, 3, 4, 5 & 1, 6, 7 & -- & 0.53 & 4,777 \\
    \hline
  \end{tabular}
  \caption{From 7 categories to binary claim vs ``not a claim'' classification. N = number of sentences annotated by majority. (1) Personal experience, (2) Quantity in the past or present, (3) Correlation or causation, (4) Current laws or rules of operation, (5) Prediction, (6) Other type of claim, (7) Not a claim.}
  \label{tab:reformulations}
\end{table}

For mapping the 7 categories in the schema into the binary classification task distinguishing claims and non-claims, two different mappings were initially proposed (see Table \ref{tab:reformulations}). These two methods were proposed by first assuming the two extremes, i.e. Quantity (2) should be deemed a claim, whereas Not a claim (7) belong to non-claim. Subsequently, these two reformulations of the taxonomy were proposed after meeting and brainstorming with factcheckers. While the method in the first row would be reasonable for achieving higher inter-annotator agreement, it would lead to a classification performance prioritising high precision at the expense of a lower recall. The method in the second row was ultimately selected for further experimentation, as recall was deemed important by factcheckers and it is in turn more realistic for not omitting any of the 7 initial categories.

Hence, we moved on to evaluate our claim detection system on the labels originating from the second row of Table \ref{tab:reformulations}. Here `Other type of claim' is not in the positive class for two reasons. First of all there is a lot of disagreement between it and the `Not a claim' class. Secondly, the kinds of claim in the 'Other' section - voting records, quotes, statements about public opinion polls, are less frequently written about by Full Fact.

The agreement of 60\% is still low -- that is a lot of sentences to throw away if we were only to consider agreement among all the annotators. So instead we choose a majority vote where at least 3 annotators marked the sentence and more than half of them agree. Out of the initial 6,304 sentences this filter selects 4,777 sentences, 3,973 not claims and 804 claims. This is in line with previous studies where the proportion of claims is 10-30\% in political TV \cite{hassan2017claimbuster,Gencheva2017ACA}. As extra training data we add 794 claims from the Full Fact database. Out of them 766 are annotated by us as positive because they fall into our claim categories, for example ``The courts have said that the so-called `bedroom tax' is illegal.'' The remaining 28 are in the `Other type of claim' category, for example ``The British economy is not only getting better, it is healing.''

If we keep the 7 categories in the dataset, instead of mapping them into the 2 classes, the same method based on majority votes leads to a slightly smaller dataset with 4,080 sentences, i.e. due to the slightly lower agreement on the broader set of 7 categories.

\section{Methods}

To capture the diversity of sentences observed during political TV shows, we propose to leverage universal sentence representations. We use InferSent \cite{conneau2017supervised} as a method to achieve sentence embeddings. These embeddings are different from averaging word embeddings because they take word order into account using a recurrent neural network. The method provided by InferSent involves words being converted to their common crawl GloVe implementations before being passed through a bidirectional long-short-term memory (BiLSTM) network \cite{hochreiter1997long}. The sentence embeddings were pre-trained on a large dataset of Natural Language Inference tasks\footnote{https://nlp.stanford.edu/projects/snli/}. Additionally, we also tried concatenating POS and NER information to the embeddings. For each sentence, the POS/NER feature vector was the count of each POS/NER tag in the corpus. We input our sentence representations to a range of supervised classifiers implemented using scikit-learn \cite{scikit-learn}, with the classifiers set to their default parameters. The four classifiers we tested include Logistic Regression, Linear SVM, Gaussian Na\"ive Bayes and Random Forests, all of which use the default parameters provided with scikit-learn.

We use a number of other features as baselines:

\begin{enumerate}
    \item A number of variants of the state-of-the-art claim detection system by ClaimBuster, using different combinations of TF-IDF, POS and NER features, as in \cite{hassan2017claimbuster}.
    \item Averaging pre-trained word embedding vectors for all words in a sentence. We evaluate:
    \begin{itemize}
        \item Word2vec \cite{mikolov2013efficient} via the Gensim implementation \cite{rehurek_lrec}, using the \textit{GoogleNews} embedding.
        \item GloVe \cite{pennington2014glove} trained on Common Crawl, as well as combining them with dimensionality reduction using principal component analysis (PCA).
    \end{itemize}
    \item TF-IDF representations of sentences with logistic regression. Numbers have a significant role in claims - the ``Cardinal Number'' part-of-speech tag is the second most discriminating feature in \cite{hassan2017toward}, so we try a Spacy NER to replace numbers with `*NUMBER*' during preprocessing.
\end{enumerate}

For our implementation of the ClaimBuster system, we use the Watson Natural Language Understanding API, the updated version of the Alchemy API used by the original authors. All other features were implemented as outlined in \cite{hassan2017claimbuster}.

The ClaimRank \cite{Gencheva2017ACA} model was harder to re-implement. In order to maintain impartiality Full Fact cannot use the data on sentence speakers when selecting which claims to check. This special care is also encouraged in the computer science research for systems that are integrated into the infrastructure of society by the ACM code of ethics \cite{ACMEthics}. Additionally, our dataset didn't have data on applause, laughing, or speaker crossover. Even though we unfortunately couldn't use one third of ClaimRank's features\footnote{https://github.com/pgencheva/claim-rank}, we trained the FNN, SVM and logistic regression classifiers on the remaining features in our dataset \cite{Gencheva2017ACA}.\footnote{Note that ClaimRank is originally a ranking system that determines the check-worthiness of claims, so here instead we implemented several binary classifiers to make it work as a binary claim detection system.}

\subsection{Experiment Settings}

The dataset consists of 5,571 sentences (4,777 from annotations and 794 from the Full Fact database of claims), of which 1,570 are claims and 4,001 are not claims, which gives a 30/70 class imbalance. We use stratified 5-fold cross-validation to train and test our models. We use precision, recall and F$_1$-score measures to assess classifier performance. 

We show the best-performing classifier for any given feature set. We also show 95\% confidence interval for the precision and recall using binomial distributions. This demonstrates possible overlap in results between different models. The interval is wide for recall due to the small number of positive examples. The next section will present the results of applying these methods.

\section{Claim Detection}

Here we present results for the binary classification, using the class grouping shown in the second row of Table \ref{tab:reformulations}.

\subsection{Analysis of Results}

Table \ref{tab:results} shows the results of our experiments. Note that in this task recall is especially important as factcheckers do not want to miss out important claims, and hence our priority is to maximise recall while also keeping a good balance of precision and recall as measured by the F1 score. Interestingly, the simple approach of TF-IDF achieves high precision but low recall. We call our new model `CNC' which stands for ``Claim/No Claim''. It achieves a better balance of precision and recall; logistic regression classifier gives the highest overall F1 score of 0.83, outperforming all other techniques. In the interest of space and clarity, we do not include the results of the other classifiers (i.e. Linear SVM, Gaussian Na\"ive Bayes and Random Forests). The use of POS and NER features in our model has no effect on the performance. GloVe embeddings achieve performance close to our method with F1 scores 2\% lower and substantially lower recall scores. Despite the overlap in precision scores between GloVe and our method, the overlap is minimal in terms of recall. Our CNC model also clearly outperforms the state-of-the-art method by ClaimBuster at 0.79 F1; hence our method yields F1 scores that improve ClaimBuster by over 5\% in relative terms. ClaimBuster performs similarly to CNC in terms of precision, albeit with substantially lower recall scores. ClaimRank has the best precision scores across the board, but with the lower recall scores. CNC achieves a 6\% relative improvement in F1-score over ClaimRank. 

\begin{table}[htb]
  \centering
  \begin{tabular}{|l|l||c|c|c||c|c|}
    \hline
    \textbf{Features} & \textbf{Classifier} & \textbf{P} & \textbf{R} & \textbf{F$_1$} & \textbf{P-interval} & \textbf{R-interval} \\
    \hline
    \hline
    TF-IDF & LogReg & .90 & .59 & .70 & .89 - .91 & .56 - .61\\
    \hline
%    TF-IDF + regex number preproc. & LogReg & \textbf{.91} & .59 & .70 & .90 - .92 & .56 - .61\\
%    \hline
    TF-IDF+num. preproc. & LogReg & .91 & .59 & .70 & .90 - .92 & .56 - .61\\
    \hline
    \hline
    Word2Vec & SVM & .85 & .75 & .78 & .84 - .86 & .73 - .77 \\ 
    % \hline 
    % GloVe  & SVM & .87 & .78 & .81 & .86 - .88 & .76 - .80 \\
    \hline
    GloVe & LogReg & .89 & .76 & .81 & \textbf{.88 - .90} & \textbf{ .74 - .78} \\
    \hline
    GloVe+PCA & LogReg & .89 & .75 & .81 & .88 - .90 & .73 - .77 \\
    \hline
    \hline
    CB & LogReg & .90 & .59 & .70 & .89 - .91 & .56 - .61  \\
    \hline 
    CB +POS & LogReg & .88 & .68 & .76 & .86 - .89 & .66 - .71 \\
    \hline 
    CB +NER & LogReg & .88 & .60 & .71 & .87 - .89 & .58 - .63 \\
    \hline 
    CB +POS+NER & LogReg & .87 & .71 & .78 & .86 - .88 & .68 - .73 \\
    \hline 
    CB & SVM & .84 & .70 & .76 & .83 - .85 & .69 - .73 \\
    \hline 
    CB +POS & SVM & .86 & .74 & .79 & .85 - .87 & .72 -.76 \\
    \hline 
    CB +NER & SVM & .84 & .71 & .77 & .83 - .85 & .69 - .73 \\
    \hline 
    CB +POS+NER & SVM & .86 & .75 & .79 & .85 - .87 & .73 - .77\\
    \hline
    \hline 
    ClaimRank & LogReg & \textbf{.93} & .65 & .77 & .92 - .94 & .63 - .67 \\
    \hline 
    ClaimRank & SVM & \textbf{.93} & .53 & .67 & .92 - .94 & .51 - .55 \\
    \hline 
    ClaimRank & FNN & .89 & .61 & .72 & .87 - .91 & .58 - .62 \\
    \hline
    \hline
    \textbf{CNC} & LogReg & .88 & \textbf{.80} & \textbf{.83} & \textbf{.87 - .89} & \textbf{.78 - .82} \\ 
    \hline
    \textbf{CNC+POS} & LogReg &  .88 & \textbf{.80} & \textbf{.83} & .87 - .89 & .78 - .82 \\ 
    \hline
    \textbf{CNC+NER} & LogReg &  .88 & \textbf{.80} & \textbf{.83} & .87 - .89 & .78 - .82 \\ 
    \hline
    \textbf{CNC+POS+NER} & LogReg & .88 & \textbf{.80} & \textbf{.83} & .87 - .89 & .78 - .82 \\ 
    \hline
  \end{tabular}
  \caption{Results for the Claim/No claim experiments. CB: ClaimBuster. The 95\% confidence intervals are from binomial distribution adapting \protect\cite{kohavi1995study}.}
  \label{tab:results}
\end{table}

\section{Multi-class Classification}

We further test our classifier on the broader set of 7 categories, using the subset of 4,080 sentences where there was enough agreement at this level of granularity. We train a multinomial logistic regression on the features from CNC, the best performing binary classifier.

\subsection{Analysis of Results}

\begin{table}[htb]
  \centering
  \begin{tabular}{|l|c|c|c|c|c|c|}
    \hline
    \textbf{Class} & \textbf{P} & \textbf{R} & \textbf{F1} & \textbf{N} \\
    \hline
    \hline
    Not a claim               & .77      & .90      & .83       & 2235 \\
    \hline
    Other type of claim       & .59      & .55      & .57       & 952  \\
    \hline
    Quantity (past/present)   & .80      & .79      & .79       & 403  \\
    \hline
    Prediction                & .60      & .27      & .37       & 181  \\
    \hline
    Personal experience       & .72      & .39      & .50       & 124  \\
    \hline
    Correlation/causation     & .50      & .13      & .21       & 107  \\
    \hline
    Current laws/rules        & .27      & .04      & .07       & 78   \\
    \hline
    \hline
    \textbf{microavg / total} & .71      & .73      & .70       & 4080 \\
    \hline
    \textbf{macroavg / total} & .61      & .44      & .48       & 4080 \\
    \hline
  \end{tabular}
  \caption{Multi-class classifier performance. CNC model.}
  \label{tab:results7}
\end{table}

Table \ref{tab:results7} shows the results for the multi-class classification experiments. These results reaffirm our expectations that, beyond the binary classification of claims and non-claims, classification at a finer granularity becomes more challenging. This is especially true for the categories with the smallest number of instances, such as ``Current laws'' or ``Correlation or causation''. We achieve low F1 score for these categories, however the small number of instances may have a significant impact on this. Looking at bigger classes, ``Quantity'' (relatively easy to identify by looking for numbers and quantitative words) and ``Not a claim'' (the most popular category) yield the best $F_1$-scores. The results for ``Other type of claim'' are also good, which only tend to be confused with ``Not a claim'' -- this is in line with human annotator disagreement in Figure \ref{fig:annotation}. Overall results are reasonably good when we measure with a microaveraged F1 score of 0.70, however it shows significant room for improvement when we measure it by macroaveraged F1 score of 0.48. We aim to expand our dataset in the near future to circumvent these issues.

\begin{figure}[htb]
\centering
\includegraphics[width=0.9\textwidth]{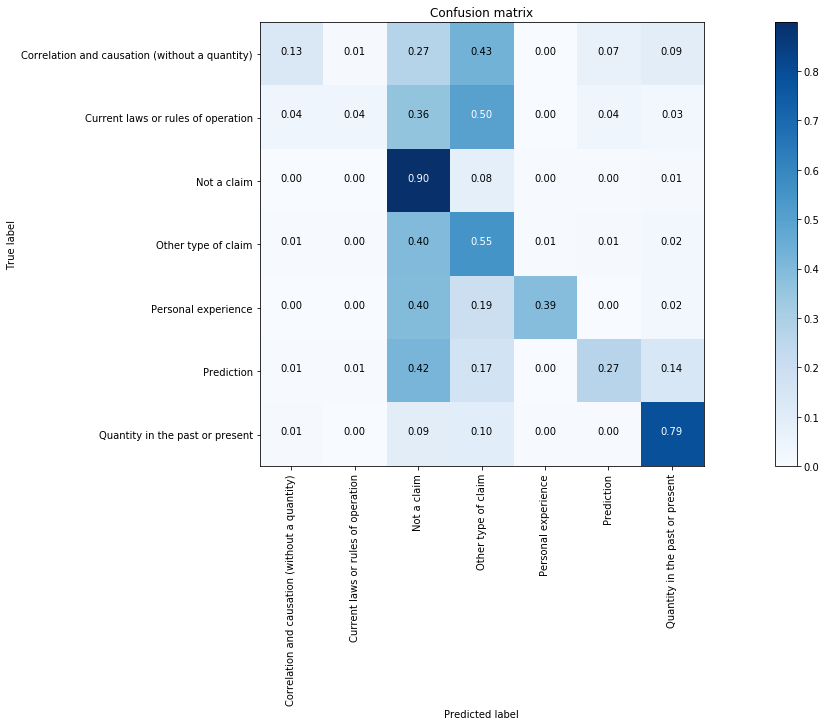}.
  \caption{Confusion matrix for the multi-class classifier.}
  \label{fig:confusion}
\end{figure}

Figure \ref{fig:confusion} shows a confusion matrix of the predictions made by the multi-class classifier. This again corroborates that ``Quantity'' is the easiest category to predict, with very few mistakes deviating to that category. Categories where more mistakes occur include ``Not a claim'' and ``Other type of claim'', which the classifier has a tendency to mispredict when the actual label is a different one. Collection of more annotated sentences is expected to yield improved performance on these categories, however the binary classifier is for now more accurate and realistically usable.

\section{Deployment and Impact}

Model deployment has impacted Full Fact and the crowdsourcing exercise has educated the volunteer community. 

Full Fact's factcheckers are provided with a UI that shows a live feed of transcripts from television in a tool called ``Live'' which aids live factchecking. When the claim detection model identifies a claim in these sentences, it highlights it in bold. In addition to this, the factcheckers have the ability to manually highlight (in yellow) any claim they believe to be of interest. The integration of claim detection model has provided several benefits. It has saved time  -- factcheckers can quickly skim through the text looking for claims, instead of reading the entire text. (Though, to ensure 100\% recall there are still designated people who watch the entire programme or read the entire transcript.) As a consequence, some factcheckers have started skimming just the claims in the transcripts in areas that are outside of their immediate domain. Going through the entire transcript was not viable for them prior to this automation.

We analysed 4 live factchecking sessions and noted that all claims manually highlighted in yellow had also been detected by the model, demonstrating the high recall. The precision however is harder to measure during user deployment. The model does detect claims that aren't highlighted in yellow, but this is to be expected. The reason for this is that factcheckers rely on their domain expertise and awareness of current affairs to decide which claims are of specific interest to them -- as opposed to highlighting all claims that they notice. These considerations are impossible to encode in a model as discussed in our claim definition section.

A factchecker gave the following feedback on the system: \textit{``Claim detection is very useful after I have finished live factchecking a show and reviewing it to decide what to write a longer piece about. I no longer have to read the whole transcript, just the highlighted bits.''}

Outside of the Full Fact factcheckers, there has also been anecdotal evidence of impact on the volunteer community. The annotation exercise has been educational for them. They became more scrupulous media consumers. This was also a notable side product in other crowdsourced factchecking initiatives such as TruthSquad\footnote{``Crowdsourced Fact-Checking? What We Learned from Truthsquad'' 2010 \url{https://perma.cc/J8AS-YU8E}}.

\begin{figure}[htb]
\centering
\includegraphics[width=14cm]{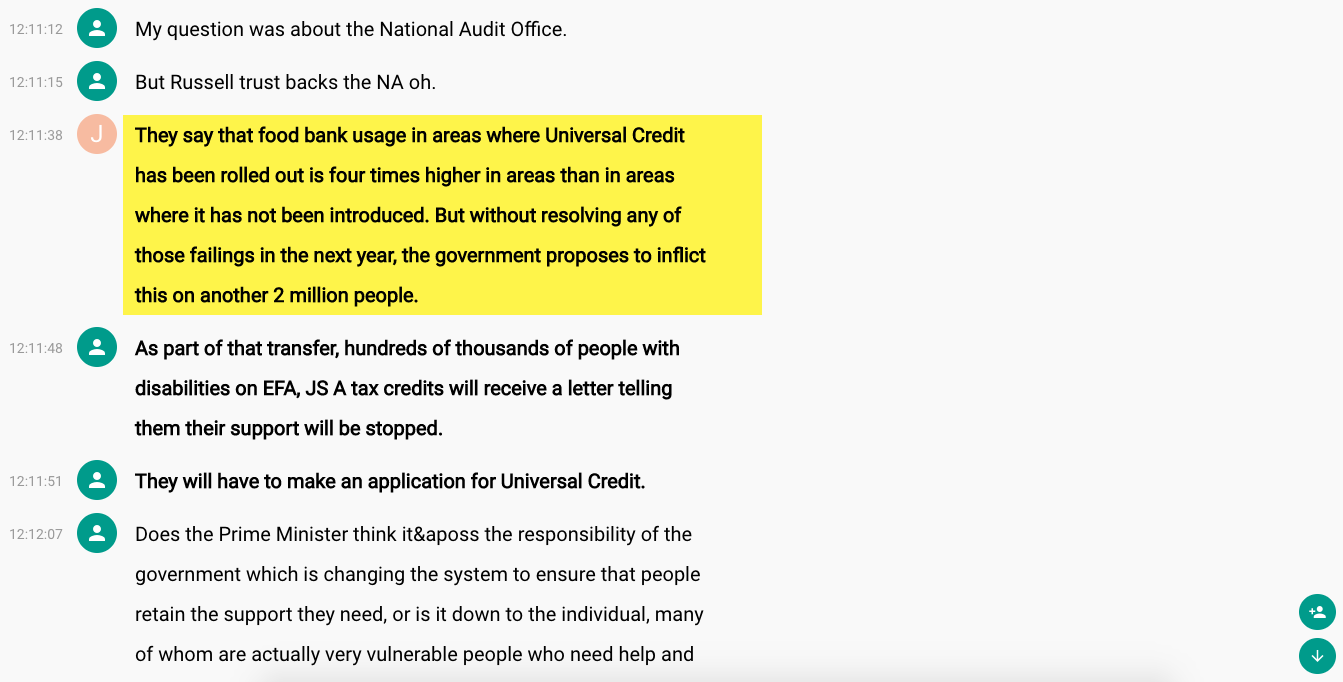}.
  \caption{Claims highlighted in bold in the ``Live'' factchecking tool during Prime Minister's Questions on 12 Sept 2018. The transcript errors are from the closed captions broadcast.}
  \label{fig:cncinaction}
\end{figure}

\section{Conclusion}

Through leveraging the factcheckers at Full Fact, and through academia-industry collaboration, we have developed the first annotation schema for claim detection informed by experts. This has enabled us to create an annotated dataset made of sentences extracted from transcripts of political TV shows. We have introduced and tested a classifier that leverages universal sentence representations which, with an F1 score of 0.83, outperforms a range of baseline classifiers, including a well-known method by ClaimBuster, by over 5\% in relative terms. While we achieve competitive results for binary claim classification, there is room for improvement when we need finer granularity of classification into the 7 categories in the annotation schema.

Our plans for future work include expanding our dataset to other languages and collaborating with other factchecking organisations. We also wish to collect more data using the same annotation schema from other news sources, such as social media, digital and print outlets. Another small potential improvement is using the counts of first/second person pronouns to detect ``Personal Experience'' category as proven useful in \cite{park2014identifying}.

\section*{Acknowledgments}

We would like to thank all 80 volunteers who participated in the annotation task, in particular Andreas Sampson Geroski. Also, thanks to Joseph O'Leary, Amy Sippitt, Will Moy, Ed Ingold, Vigdis Hanto, Mats Tostrup, Heri Ramampiaro, Jari Bakken, Benj Petitt, Michal Lopuszynski, Ines Montani, Gael Varoquaux, David Corney.

This work has been partially funded by Open Society Foundation (OR2017-35395) and Omidyar Network.

\bibliographystyle{ACM-Reference-Format}
\bibliography{claims}

\end{document}